\documentclass[twoside]{article}

\usepackage[accepted]{aistats2019}
\usepackage{graphicx}
\usepackage{subcaption}
\usepackage{url}
\usepackage{amsfonts}
\def\D{\mathrm{d}}

\usepackage{tikz}
\DeclareRobustCommand\full
    {\tikz[baseline=-0.6ex]\draw[thick] (0,0)--(0.5,0);}

\DeclareRobustCommand\dashed
    {\tikz[baseline=-0.6ex]\draw[thick,dashed] (0,0)--(0.54,0);}

\begin{document}



\twocolumn[

\aistatstitle{Approximate Inference for Multiplicative Latent Force Models}

\aistatsauthor{ Daniel J. Tait \And Bruce J. Worton}

\aistatsaddress{ School of Mathematics \\University of Edinburgh
  \And
  School of Mathematics \\University of Edinburgh} ]

\begin{abstract}
  Latent force models are a class of hybrid models for dynamic systems,
  combining simple mechanistic models with flexible Gaussian process (GP)
  perturbations. An extension of this framework to include multiplicative
  interactions between the state and GP terms allows strong a priori
  control of the model geometry at the expense of tractable inference.

  In this paper we consider two methods of carrying out inference within
  this broader class of models. The first is based on an adaptive gradient
  matching approximation, and the second is constructed around mixtures
  of local approximations to the solution. We compare the performance of
  both methods on simulated data, and also demonstrate an application of
  the multiplicative latent force model on motion capture data.
\end{abstract}

\section{Introduction}

Historically the modelling of dynamic systems broadly followed one of
two distinct philosophies; the first of these is classical and
often referred to as the \emph{mechanistic approach} which aims to
construct realistic models guided by sound principles. In contrast, the
\emph{data driven} paradigm, inspired by modern machine learning techniques,
places a greater emphasis on prediction, and allowing the observables to
guide the processes of pattern discovery. The conflict between these two
philosophies can be particularly pronounced for complex dynamic systems,
when a complete mechanistic description is often difficult to motivate, but
models with some degree of physical realism are likely to be more effective
extrapolating from the training data. Therefore, it would be desirable
to have a framework that allows for the specification of a simplistic
representation of the driving dynamics, while still allowing for relevant
dynamic systems properties to be encoded into the model. One successful
approach to constructing such hybrid models is the latent force model
(LFM) introduced in \cite{alvarez}. By combining a simple class of mechanistic
models with the flexibility offered by inhomogenous Gaussian process (GP)
pertubations one is able to construct a GP regression, with dynamic systems
properties encoded within the kernel function.

While the GP regression framework allows for tractable inference this assumption
is also one of the primary constraints of the LFM. In an effort to move beyond
this constraint \cite{tait} introduced an extension of this model, while staying
faithful to the underlying modelling framework; a linear time dependent
ordinary differential equation (ODE) in which the time dependent behaviour arises
from the variations of a set of
independent GP variables, but now allowing multiplicative interactions between the
state and latent forces. The result is a semi-parametric model that allows for
the embedding of rich topological structure.

Unfortunately, the greater control of the model geometry comes at the expense of
the tractable GP regression framework of the LFM, and therefore we must necessarily
consider approximate inference methods. In this paper
we consider two methods of introducing approximate likelihood functions for this
class of models. The first is an application of adaptive gradient matching
methods introduced by \cite{calderhead, donde} for handling problems in the very
general class of nonlinear ODEs with random parameters. The second combines
truncated local approximations to the pathwise solution of the ODE using a
mixture modelling approach.

In the next section we provide a review of the LFM framework, including the extension
allowing for multiplicative interactions. Then in Section \ref{sec:adapgrad} we
discuss how the adaptive gradient matching methods can be used to introduce an
approximate inference method for the extended of the latent force model,
and demonstrate they lead to a practical simplification in the linear case. We
then consider the method based on local successive approximations in Section \ref{sec:sa}. In Section \ref{sec:sim}
we compare the performance of the two methods on simulated data, and in Section
\ref{sec:mocap} we demonstrate the viability of the model with multiplicative
interactions with an application to the modelling of human motion capture data using
geometry constrained models before concluding with a discussion.

\section{LATENT FORCE MODELS}
\label{sec:lfm}

Latent force models \cite{alvarez} are a class of hybrid
models of dynamic systems providing a compromise between purely data
driven approaches, and more involved mechanistic models. They combine the simplest
class of mechanistic models; linear ODEs with, diagonal, constant coefficient
matrices with the flexibility of an additive GP forcing term. In what follows
we let $\mathbf{x}(t)$ denote a $K$-dimensional state variable, and we collect $R$
independent, smooth, GPs into a vector valued process
$\mathbf{g}(t) = (g_1(t),\ldots,g_R(t))^{\top}$. Then the LFM is described by the
initial value problem (IVP)

\begin{align}
  \frac{\D\mathbf{x}(t)}{\D t} = \mathbf{D}\mathbf{x}(t)
  + \mathbf{S}\mathbf{g}(t), \label{eq:lfm}
\end{align}

where $\mathbf{D}$ is a diagonal matrix, and the sensitivity matrix, $\mathbf{S}$,
is a $K \times R$ real valued rectangular matrix. From \eqref{eq:lfm} it is clear
that the only interactions between the state variables, $x_k(t)$, is through the
common latent force variables, and the sensitivity matrix governs the topology of
these interactions. The LFM encodes dynamic systems properties, but still allows for
tractable inference because the solution may be expressed as a
linear transformation of the latent GPs

\begin{align}
  \mathbf{x}(t) = e^{\mathbf{D}(t-t_0)}\mathbf{x}(t_0)
  + e^{\mathbf{D}t}\int_{t_0}^{t}
  e^{-\tau \mathbf{D}t}\mathbf{S}\mathbf{g}(\tau)
  \D\tau,
  \label{eq:lfm_sol}
\end{align}

this linear relationship between the states and latent GPs results in a
joint Gaussian distribution. This property makes it possible to marginalise
over the latent forces, so that inference for the LFM may proceed exactly
as in standard GP regression.

The GP regression framework leads to straightforward inference, but the
assumption of Gaussian trajectories of the state variable may be implausible;
this will be the case for time series of circular, directional data, and tensor
valued data. With this in mind \cite{tait} proposed an extension of the LFM
retaining the linear ODE framework, but allowing for non-Gaussian trajectories
by including multiplicative interactions between the GP forcing functions and the
state variables which they refer to as the multiplicative latent force model (MLFM).

The MLFM may be represented by the linear ODE
\begin{align}
  \frac{\D \mathbf{x}(t)}{\D t}
  = \mathbf{A}(t)\mathbf{x}(t),
  \qquad \mathbf{A}(t)
  = \mathbf{A}_0 + \sum_{r=1}^{R}\mathbf{A}_r \cdot g_r(t).
  \label{eq:mlfm}
\end{align}

The dynamics will be governed by the support of the $K\times K$ coefficient
matrix $\mathbf{A}(t)$, which by linearity will be a matrix valued GP interacting
multiplicatively with the state variable, the support of this GP will be determined
by the coefficient matrices $\mathbf{A}_r$.

To add further flexibility we allow each of the structure matrices,
$\mathbf{A}_r, r=0,1,\ldots,R$ to be given as a linear combination of a set
of shared basis matrices $\mathbf{L}_d$, $d=1,\ldots, D$, that is
\begin{align}
  \mathbf{A}_r = \sum_{d=1}^{D} \beta_{rd} \mathbf{L}_d.
\end{align}
We denote the set of connection coefficients by the $(R+1)\times D $ matrix
$\mathbf{B}$ with $\mathbf{B}_{rd} = \beta_{rd}$. Including these variables
allows a small set of forces to generate a broad range of motions,
and they play a role analogous to the sensitivity matrix in \eqref{eq:lfm}.
As is common in many modern machine learning techniques this increase in
flexibility, and predictive power, comes at the expense of identifiability
of individual parameters.

The multiplicative interactions in \eqref{eq:mlfm} and the freedom to chose
the support of the coefficient matrices enables the modeller to embed strong
geometric constraints into the pathwise solution of this model. Noteworthy 
is the case when the elements $\{\mathbf{L}_d \}_{d=1}^D$ are members of a Lie
algebra $\mathfrak{g}$ with corresponding matrix Lie group $G$, \cite{hall}.
Since $\mathfrak{g}$ is a vector space the support of $\mathbf{A}(t)$ will be
contained within this Lie algebra. It follows from this constraint that the
fundamental solution of \eqref{eq:mlfm} will itself be a member of the Lie group
$G$, \cite{iserles}, therefore allowing the construction of models either on this
group, or formed by the action of random elements within this group on some vector
space. This possibility to embed strong geometric constraints within the model is
the primary motivation for introducing this extension of the LFM.

\begin{figure}[h]
  \begin{center}
    \includegraphics[width=5cm]{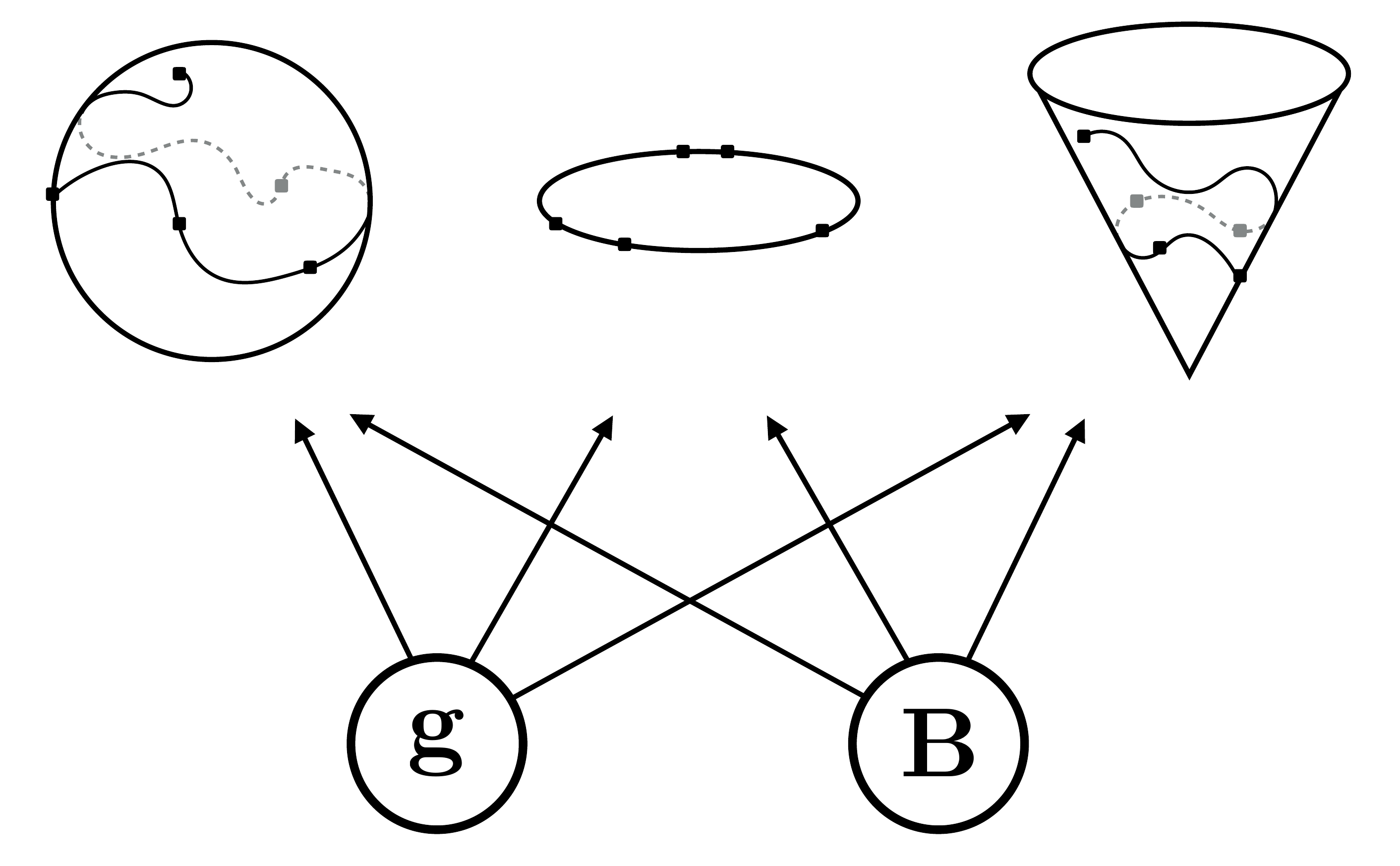}
  \end{center}
\vspace{.15in}
\caption{Topologically constrained trajectories driven by a set of common
  latent forces.}
\label{fig:mmlfm}
\end{figure}

The remarks on the geometry preservation above combined with the flexibility of choosing
the support of $\mathbf{A}(t)$ via the choice of the basis matrices $\{\mathbf{L}_d\}$
and their connection coefficients $\mathbf{B}$ allow the MLFM to provide a
straightforward conceptual framework for considering dependent processes on distinct
manifolds. In particular, we consider the case where the data space
may be factored into a collection of manifolds $\mathcal{M}_q$, $q=1,\ldots, Q$.
On each of these manifolds we assume the subprocess is modelled by
\eqref{eq:mlfm}, but with a shared set of common latent force functions. The
variations of these forces will be modulated by linear combinations of the
manifold dependent connection coefficients, $\mathbf{B}^{(q)}$. This conceptual
framework is represented graphically in Figure \ref{fig:mmlfm} where the choice of
manifold dependent basis matrices and coefficients allows for topologically distinct
trajectories driven by a set of common forces. An analogous interpretation is
available for the LFM, but in this case the product topology is just the Cartesian
product of one dimensional real spaces.

While the extension to geometrically structured, non-Gaussian, trajectories
allows for increased modelling capabilities, there is no longer a simple
closed form solution for the pathwise trajectories analogous to
\eqref{eq:lfm_sol}. Indeed, the analogous pathwise solution may be given
by the expansion
\begin{align}
  \mathbf{x}(t) 
  = \bigg(I &+ \int_{t_0}^t \mathbf{A}(\tau)\D\tau \notag \\
  &+ \int_{t_0}^{t}\mathbf{A}(\tau_1)
  \int_{t_0}^{\tau_1}\mathbf{A}(\tau_2)\D\tau_2\D\tau_1
  + \cdots\bigg)\mathbf{x}(t_0) . \label{eq:neuexpansion}
\end{align}
The presence of products of the matrix GP within each of the integrands
make it unclear as to how the distribution of the GP terms will propagate
to the state space, and unlike the case for \eqref{eq:lfm_sol} it may not
be possible to marginalise over these variables. Because we cannot perform
this marginalisation step we will, in the remainder of this paper, consider
methods for approximating the conditional distributions around a given
sample of the latent GPs.

\section{ADAPTIVE GRADIENT MATCHING}
\label{sec:adapgrad}

Bayesian adaptive gradient matching methods were introduced in
\cite{calderhead, donde} for carrying out approximate inference
of the evolution of a $K$-dimensional state variable described
by the very general class of, possibly nonlinear, ODEs

\begin{align}
  \frac{\operatorname{d} \mathbf{x}(t)}{\operatorname{d}t}
  = f(\mathbf{x} , \boldsymbol{\theta}), \label{eq:ode}
\end{align}

where the smooth function $f(\cdot ; \boldsymbol{\theta})$ is parametrised
by some random vector $\boldsymbol{\theta}$. While initially applied to the
inference of a finite dimensional parameter vector it is, in principle,
straightforward to extend the approach to the infinite dimensional case.

We shall be interested in carrying out posterior inference on the
basis of a collection of $\mathbb{R}^K$ valued random variables
$\mathbf{Y} = \{ \mathbf{y}(t_1), \ldots, \mathbf{y}(t_N) \}$ observed
at times $t_1 < \cdots < t_N$. Each of these variables is assumed to be
an independent noisy observation of the state variable, $\mathbf{x}(t)$,
the evolution of which is described by \eqref{eq:ode}. We denote the complete
collection of state variables by $\mathbf{X}$, and also consider the vectors
formed taking only the $k$th dimensional component at each time point which we
denote by $\mathbf{x}_k := (\mathbf{x}_k(t_1), \ldots, \mathbf{x}_k(t_N))^{\top}$.
If will also be convenient to define the $N$-vector $\mathbf{f}$ with
components $(\mathbf{f}_k)_i := (f(\mathbf{x}(t_i), \boldsymbol{\theta}))_k$,
$i=1,\ldots, N$.

Generally speaking, inference for ODE problems with random parameters,
is difficult because the state is only given implicitly as a transformation
of the stochastic processes. The problem would be simpler if observations
of the gradient process
were also available because in that case such a transformation is given
explicitly by \eqref{eq:ode}. Adaptive gradient matching treats
the gradient as a missing variable, and attempts to introduce an approximation
to the complete data likelihood which allows for the gradients to be
marginalised out. This is done by initially placing independent
on each component of the state variable
\begin{align}
  p_{GP}(\mathbf{x}_k) = \mathcal{N}(
  \mathbf{x}_k \mid \mathbf{0}, \mathbf{C}_{\phi_k}),
\end{align}
where $\mathbf{C}_{\phi_k}$ is the $N\times N$ covariance matrix obtained by
evaluating the kernel function of each latent state interpolating process
$k(\cdot, \cdot ; \phi_k)$. Each GP is assumed differentiable allowing the
construction of a conditional distribution for the gradients, $\dot{\mathbf{x}}_k$,
under the prior which we denote by
\begin{align}
  p_{GP}(\dot{\mathbf{x}}_k \mid \mathbf{x}_k) :=
  \mathcal{N}(\mathbf{m}_{\dot{x}_k|x_k}, \mathbf{C}_{\dot{x}_k\mid x_k}),
  \label{eq:pxdot_gp}
\end{align}
where the conditional mean and conditional covariance matrices are obtained
from the joint Gaussian distribution of the state and its gradient \cite{solak}.

The conditional distribution \eqref{eq:pxdot_gp} contains no reference to the
model \eqref{eq:ode}, and therefore \cite{calderhead} consider introducing a
separate conditional density with the nonlinear regression form

\begin{align}
  p_{reg} (\dot{\mathbf{x}}_k \mid \mathbf{X}, \boldsymbol{\theta}, \gamma_k) =
  \mathcal{N}(\mathbf{f}_k(\mathbf{X}, \boldsymbol{\theta}), \gamma \mathbf{I}_N ),
  \label{eq:pxdot_reg}
\end{align}

where $\gamma_k$ represents a temperature parameter controlling the extent to
which the conditional distribution is constrained by the functional form of
the model.

The two dissonant conditional distributions \eqref{eq:pxdot_gp} and
\eqref{eq:pxdot_reg} are then combined to form a single distribution
using a product of experts approximation

\begin{align}
  p(\dot{\mathbf{X}}\mid \mathbf{X}, \boldsymbol{\theta})
  &\propto
  \prod_{k=1}^K p_{GP}(\dot{\mathbf{x}}_k \mid \mathbf{x})
  p_{reg}(\dot{\mathbf{x}}_k \mid
  \mathbf{f}_k(\mathbf{X}, \boldsymbol{\theta}), \gamma_k\mathbf{I}_N).
  \label{eq:prodexp}
\end{align}

The result is a conditional density which places most of its mass around estimates
of the gradient which agree with the parametrised model \eqref{eq:ode}, but also
coincide with the gradient of a GP interpolant.

Since this distribution is a product of Gaussians it is possible to marginalise
over the gradients, \cite{donde}, leading to
\begin{align}
  p(\mathbf{X} \mid \boldsymbol{\theta})
  &= \int p(\dot{\mathbf{X}} , \mathbf{X} \mid \boldsymbol{\theta} )
  \D\mathbf{X}\notag \\ 
  &\propto
  \exp\bigg\{
  -\frac{1}{2}\sum_{k=1}^{K}
  \boldsymbol{\eta}_k^{\top}
  (\mathbf{C}_{\dot{x}_k|x_k} + \gamma_k I)^{-1}
  \boldsymbol{\eta}_k \notag\\
  &\qquad{}\qquad -\frac{1}{2}\mathbf{x}_k^{\top}\mathbf{C}_{\phi_k}^{-1}\mathbf{x} \label{eq:adapgradlik}
  \bigg\},
\end{align}
where we have defined
\begin{align}
  \boldsymbol{\eta}_k &=  \mathbf{f}_k - \mathbf{m}_{\dot{x}_k|x_k}. \notag
\end{align}
It is clear that if the evolution equation is a degree $P$ polynomial
in the state then the argument of the exponential in \eqref{eq:adapgradlik}
will be a degree $2P$ polynomial. Here we consider the linear case
which leads to exponential quadratics, and so tractable Gaussian posterior
conditionals.

When the evolution equation \eqref{eq:ode} is given
by the MLFM \eqref{eq:mlfm} we identify the arbitrary parameter
$\boldsymbol{\theta}$ with the set of latent forces
$\mathbf{g} = \{\mathbf{g}_1,\ldots,\mathbf{g}_R\}$ and coefficient matrix
$\mathbf{B}$. We may rewrite the variable $\mathbf{f}_k$ appearing in
\eqref{eq:adapgradlik} using the equivalent linear representations

\begin{align}
  \mathbf{f}_k = \sum_{j=1}^{K} \mathbf{u}_{kj} \circ \mathbf{x}_j
  = \sum_{r=0}^{R} \mathbf{v}_{kr} \circ \mathbf{g}_r
  = \sum_{r=0}^{R}\sum_{d=1}^{D} \beta_{rd} \mathbf{w}_{rd}, \label{eq:repr}
\end{align}

where $\circ$ denotes the elementwise product of two arrays of conforming
shape, and we define the $N$-vectors
\begin{align*}
  \mathbf{u}_{kj} = \sum_{r=0}^R A_{rkj} \mathbf{g}_r,
  \qquad \mathbf{v}_{kr} = \sum_{j=1}^{K} A_{rkj} \mathbf{x}_j,
\end{align*}
and
\begin{align*}
  \mathbf{w}_{rd} = \mathbf{g}_r \circ \sum_{d=1}^{D} \sum_{j=1}^{K} \mathbf{L}_{dkj} \mathbf{x}_j.
\end{align*}

Having defined these variables we can produce a posterior for any choice of
of the variables $\mathbf{X}, \mathbf{g}, \mathbf{B}$ by conditioning
on $\mathbf{u}$, $\mathbf{v}$ or $\mathbf{w}$ respectively, and then rearranging
the exponential quadratic accordingly. It follows that if we also place a Gaussian
prior on the coefficient $\mathbf{B}$, we will have a collection of Gaussian
posteriors
\begin{subequations}
\begin{align}
  p(\mathbf{X} \mid \mathbf{B}, \mathbf{g}) &=
  \mathcal{N}\left( \mathbf{X} \mid
  \mathbf{0},
  \mathbf{S}_x(\mathbf{B},\mathbf{g})
  \right), \\
  p(\mathbf{g} \mid \mathbf{X}, \mathbf{B}) &= \mathcal{N}\left(
  \mathbf{g} \mid
  \mathbf{m}_g(\mathbf{X}, \mathbf{B}),
  \mathbf{S}_g(\mathbf{B}, \mathbf{g})
  \right), \label{eq:ag_gpost} \\
  p(\mathbf{B} \mid \mathbf{X}, \mathbf{g}) &= \mathcal{N}\left(
  \mathbf{B} \mid
  \mathbf{m}_{\beta}(\mathbf{X}, \mathbf{g}),
  S_{\beta}(\mathbf{X}, \mathbf{g}
  \right).   
\end{align}
\end{subequations}
These closed form solutions are in contrast to the general case in which
it is necessary to sample from the density \eqref{eq:adapgradlik} with 
unknown normalising constant.

As an example $S_g(\mathbf{X}, \mathbf{B})$ is formed by inverting the $R \times R$
block matrix given by
\begin{align*}
  S_g(\mathbf{X}, \mathbf{B})^{-1}_{rs} =
  \sum_{k=1}^{K} \mathbf{v}_{kr}\mathbf{v}_{ks}^{\top}
  \circ \left(\mathbf{C}_{\dot{x}_k|\dot{x}} + \gamma_k I\right)^{-1} + \delta_{rs}
  \mathbf{C}_{\psi_r}^{-1},
\end{align*}
for $r, s = 1,\ldots, R$, and where $\mathbf{C}_{\psi_r}$ is the covariance matrix of
the $r$th latent force. Similarly, the mean is given by
\begin{align}
  \mathbf{m}_g = \mathbf{S}_g(\mathbf{X}, \mathbf{B})^{-1}\sum_{k=1}^{K} \mathbf{V}_k^{\top}(
  \mathbf{C}_{\dot{x}_k|x_k} + \gamma_k I)^{-1}(\mathbf{m}_k - \mathbf{v}_{k, 0}).
  \label{eq:mlfmag_map_g}
\end{align}

The construction of the Gaussian conditionals for $\mathbf{B}$ and $\mathbf{X}$ proceeds
in a similar way. The fact that each of the variables of principal interest is
conditionally Gaussian allows for straightforward implementation of Gibbs sampling methods,
or of mean field variational updates.

\section{MIXTURES OF SUCCESSIVE APPROXIMATIONS}
\label{sec:sa}
The key component of the adaptive gradient matching method described in the previous section
is the imposition of an approximate fixed point condition for the linear differential operator
\begin{align*}
  \mathcal{D} f(t) \stackrel{\Delta}{=} \frac{\D f(t)}{\D t} - \mathbf{A}(t)f(t),
\end{align*}
for differentiable $\mathbb{R}^K$ valued functions $f$. This fixed point condition was
encoded into the likelihood function through the gradient expert \eqref{eq:prodexp}.
In this section we consider an alternative approach by first integrating the
differential equation, and then considering fixed points of the integral operator
\begin{align}
  \mathcal{L} f(t) \stackrel{\Delta}{=} f(t_0)
  + \int_{t_0}^t \mathbf{A}(\tau)f(\tau)\D\tau.
\end{align}
We follow the construction in \cite{tait} by considering an approximation to the
distribution of the state variable $\mathbf{x}(t)$ conditional on dense sample
paths of the coefficient matrix $\mathbf{A}(t)$. The solution will be constructed
from a given initial approximation $\mathbf{x}^{(0)}$ by means of the Picard iteration
\begin{align}
  \mathbf{x}^{(n+1)}(t) = \mathbf{x}^{(n)}(t_0) + \mathcal{K}[\mathbf{x}^{(n)}](t),
  \label{eq:picard}
\end{align}
where the integral operator is defined by
\begin{align}
  \mathcal{K}[f](t) \stackrel{\Delta}{=} \int_{t_0}^t\mathbf{A}(\tau)f(\tau)\D\tau.
  \label{eq:K_integral}
\end{align}
Repeatedly iterating the map \eqref{eq:picard} from a constant initial approximation,
and then collecting terms leads to the expansion \eqref{eq:neuexpansion}. This approach
is referred to as the method of successive approximations, and is an important
construction in the classical existence and uniqueness theorems for ODEs.

In practice we have access to only a finite realisation of the sample path of the
coefficient matrix, although we may make this path arbitrarily fine at a corresponding
increase in computational complexity. Therefore assuming a suitably fine approximation
we may replace the integral \eqref{eq:K_integral} with a numerical quadrature 
\begin{align}
  \mathcal{K}[\mathbf{x}](t_n) \approx 
  \sum_{i=1}^{N} w_{ni} \mathbf{A}(t_i)\mathbf{x}_i,
\end{align}
for appropriate choice of quadrature weights $w_{ni}$. Then a discretisation of the operator
$\mathcal{K}$ is given by
\begin{align}
  \mathbf{K}[\mathbf{g}, \mathbf{B}] &=
  \sum_{n=0}^N\mathbf{w}_n\mathbf{e}_n^{T}\otimes \mathbf{A}(t_n) \notag \\
    &= \sum_{n=0}^{N}\left[\sum_{d=1}^D (\beta_{0d} + \sum_{r=1}^R g_{rn}\beta_{rd})
    \cdot \mathbf{w}_n\mathbf{e}_n^{\top}\otimes \mathbf{L}_d\right].
  \label{eq:K_integral_discrete}
\end{align}
The corresponding discretisation of the transformation \eqref{eq:picard},
which we shall denote by $\mathbf{P}_\nu = \mathbf{P}_{\nu}[\mathbf{g},\mathbf{B}]$,
may be formed by choosing an index $\nu \in \{1,\ldots,N\}$ which will correspond
to the initial time. The $K$-vector $\mathbf{x}(\tau_\nu)$ is invariant under
\eqref{eq:picard} and so we may consider the discrete approximation to this
transformation which will act on block vectors of the form
$\mathbf{v} = (\mathbf{v}_1, \ldots, \mathbf{v}_N)^{\top}$ with
$\mathbf{v}_i \in \mathbb{R}^K$ for $i=1, \ldots, N$, by the matrix/vector
operation
\begin{align}
  \mathbf{P} \mathbf{v} = \mathbf{v}_\nu \otimes \mathbf{1}_N +
  \mathbf{K}[\mathbf{g}, \boldsymbol{\beta}] \mathbf{v}.
\end{align}

As a single iteration of the map \eqref{eq:picard} only produces a first
order approximation to the solution around $\tau_\nu$ it is necessary to consider
higher order approximations by taking $M$ iterates of this map, $M \geq 1$. The
result is a discrete approximation to an $M$th order truncation of the Neumann
series expansion given by \eqref{eq:neuexpansion}. This construction allows us to
construct approximations to the pathwise solution of the state variable for given
latent parameters which will locally solve the IVP around $\tau_{\nu}$. Our
approach is to use this local approximation to motivate a regression model for
the conditional density of the state variables which takes the form
\begin{align}
  p_{\nu}(\mathbf{X}\mid\mathbf{g},\mathbf{B}, \alpha, M) =
  \prod_{n=1}^{N}\mathcal{N}(\mathbf{x}(t_n)\mid
  \mathbf{m}_{\nu, n},
  \alpha^{-1}\mathbf{I}_K),
  \label{eq:local_density}
\end{align}
where $\mathbf{m}_{\nu, n}$ is the $nth$ subvector of the expression
\begin{align*}
  \mathbf{m}_{\nu} = \mathbf{P}_{\nu}^M
  (\boldsymbol{\mu}_\nu \otimes \mathbf{1}_N).
\end{align*}

Conditional on $\mathbf{g}$ and $\mathbf{B}$, the density function
\eqref{eq:local_density} may be viewed as an $M$th order approximation to
the IVP \eqref{eq:mlfm} with initial condition $\boldsymbol{\mu}_\nu$ and
precision $\alpha$. The mean function $\mathbf{m}_{\nu}$ will be a degree
$M$ polynomial in the values of the latent forces and connection coefficients.
The parameters $\boldsymbol{\mu}_{\nu}$ are to be interpreted as initial conditions
of a local version of the IVP \eqref{eq:mlfm}, and for high precision $\alpha$
are well informed by the data.

As the construction is local in character, larger time intervals will necessarily
require higher orders of approximation, and therefore an increasing computational
burden. In order to alleviate this complexity rather than model the whole
interval with one conditional density, our approach is to pick a set of initial
times $\tau_{\nu}, \nu=1,\ldots, D$, and consider the mixture of local
regression models

\begin{align}
  p(\mathbf{X} \mid \mathbf{g}, \boldsymbol{\beta}, \alpha) =
  \sum_{\nu=1}^{D} \pi_{\nu} 
  p_\nu(\mathbf{X} \mid \mathbf{g}, \boldsymbol{\beta}, \alpha).
  \label{eq:lik_mlfmsamix}
\end{align}

Each mixture component represents a local approximation to the
solution of the MLFM around $\tau_{\nu}$. We refer to approaches using the
likelihood term \eqref{eq:lik_mlfmsamix} as MLFM by mixtures of successive
approximations (MLFM-MixSA) methods. After introducing priors for the
latent variables it is straightforward to construct maximum a posteriori
(MAP) estimates using the standard Expectation-Maximisation (EM)
approach for Gaussian mixture models. The classical EM approach to
fitting mixtures of Gaussians introduces the responsibility variables
and in this instance will act to discover regions over which an $M$th
order truncation of the expansion \eqref{eq:neuexpansion} gives a
good approximation to the solution.

\section{SIMULATION STUDY}
\label{sec:sim}
The construction of the MLFM-AG using interpolation of the state variables,
and the construction of the MLFM-MixSA using local approximations suggests
the need to investigate these methods under two regimes;
the first is the effect of the spacing $\Delta t$ between observations, and
the second the impact of the interval length each mixture component is
accounting for. For the studies in this section we investigate both
of these regimes while holding the total sample size constant, for each
experimental setting we simulate a total of $100$ experiments and report
the average error.

\subsection{Kubo oscillator}
For models in which the Lie algebra is trivial it is possible to
motivate a MAP estimate of the LF under an approximation to the
true posterior. One such case is the random harmonic oscillator,
\cite{vankampen} defined as the complex-valued ODE

\begin{align}
  \dot{z}(t) = -i g(t) z(t), \qquad z(t_0) \in S^1 \subset \mathbb{C}.
\end{align}

We use the methods introduced in this paper to construct MAP
estimates of the latent forces which we compare with the estimates
obtained using an approximation to the true conditional density
considered in \cite{tait}. For the MLFM-MixSA we fix the approximation
order at $M=5$ and consider the effiect of varying the number of
mixtures with equally spaced initial times $\tau_{\nu},$
$\nu=1,\ldots D$ with $D \in \{1, 2, 3\}$.

The results are displayed in Table \ref{tab:kubo} and
show that the adaptive gradient matching methods perform very well when
the sampling frequency is high, but the performance deteriorates
as $\Delta t$ decreases. The ability to increase the number of mixture
centers in the MLFM-MixSA model allows this method to better deal with
the case when the data is sparse relative to the system complexity.

\begin{figure}
  \centering
  \begin{subfigure}[b]{.21\textwidth}
    \includegraphics[width=\textwidth,
    trim={0 .1cm 0 0}, clip]{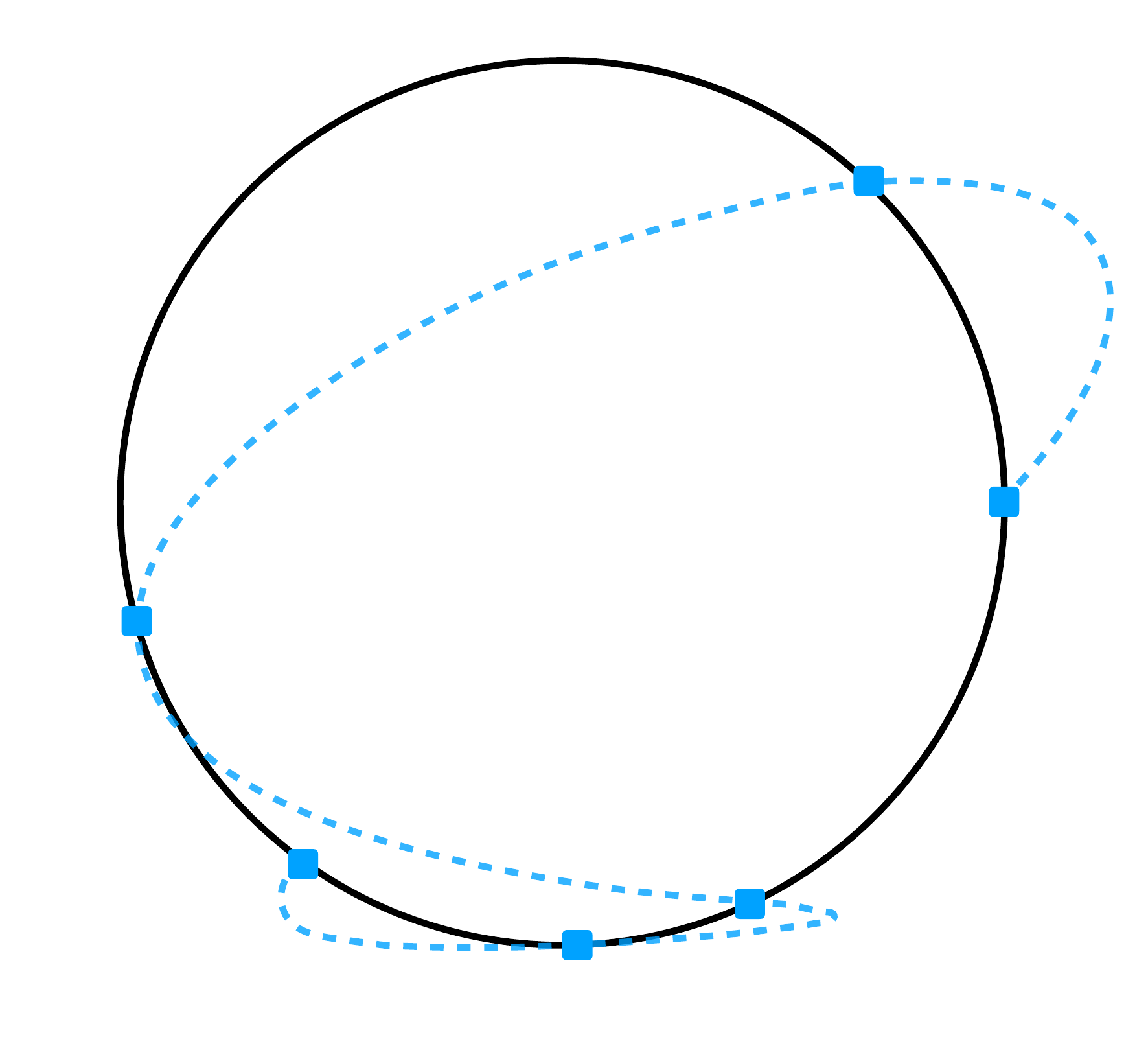}
    \caption{GP interpolant}
    \label{fig:ag_S1_a}
  \end{subfigure}
  \qquad
  \begin{subfigure}[b]{.21\textwidth}
    \includegraphics[width=\textwidth]{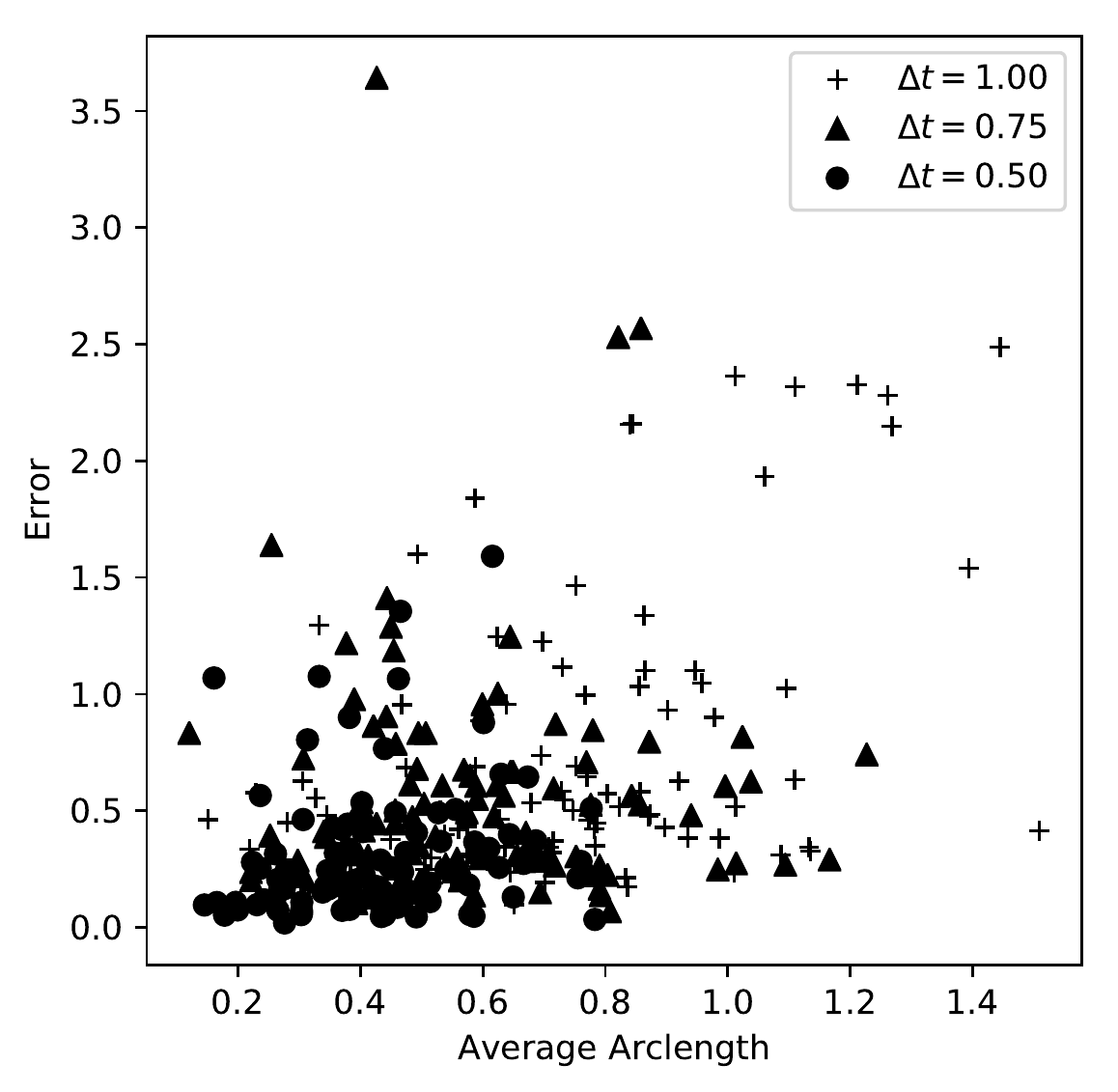}
    \caption{Error vs. arc length}
    \label{fig:ag_S1_b}
  \end{subfigure}
  \caption{(a) GP functions on $S^1$ perform poorly as distance between
    points increases. (b) The accuracy of the MLFM-AG
    approaches depends on the interpolating process, and so diminishes with
    the average arc length between points.}
  \label{fig:ag_S1}
\end{figure}

Because the MLFM-AG method is constructed around interpolating functions
it is not the frequency itself that leads to the decrease in performance,
but the likely increase in distance between points as the time between
them increases and a corresponding information loss in the interpolating
functions. The loss of structural information for GP interpolants on
manifolds is displayed in Figure \ref{fig:ag_S1_a}. In Figure
\ref{fig:ag_S1_b} we see that as the average arc length between data
points increases the performance of the MLFM-AG methods deteriorate.

\begin{table}[h]
  \caption{$\| \cdot \|_{2}$ distance of the latent force MAP estimates
    from the ground truth value on the interval $[0, 6]$. Reported are
    the results using the MLFM-AG and order $M=5$ MLFM-MixSA approximation.
  }
  \label{tab:kubo}
  \begin{center}
    \begin{tabular}{ccccc}
      $\Delta t$ & MLFM-AG & \multicolumn{3}{c}{MLFM-MixSA} \\
      & & $D=1$ & $D=2$ & $D=3$\\
      \hline \\
      $0.50 $ & 0.237 & 2.128 & 0.449 & 0.319 \\
      $0.75 $ & 0.402 & 2.006 & 0.489 & 0.410 \\
      $1.00 $ & 0.640 & 1.816 & 0.575 & 0.528
    \end{tabular}
  \end{center}
\end{table}

\subsection{Dynamic systems on SO(3)}
The MLFM framework enables the modelling of the action of a group on a
vector space, important to this process will be the ability to learn the
coefficient matrix of the IVP corresponding to the fundamental solution
\begin{align}
  \frac{\D\mathbf{x}(t)}{\D t} =
  \mathbf{A}(t) \mathbf{x}(t), \qquad \mathbf{x}(t_0) = \mathbf{I}_K.
  \label{eq:sim_example_2}
\end{align}
For this example we consider the case $\mathbf{A}(t) \in \mathfrak{so}(3)$,
the Lie algebra of infinitesimal rotations of $\mathbb{R}^3$. 

The skew-symmetry conditions of the matrix $\mathbf{A}(t)$ imply there
are only three independent components  $\{a_{12}(t), a_{13}(t), a_{2,3}(t)\}$,
the remaining being fixed by the skew-symmetry condition. Therefore we allow
$\mathbf{B}$ and $\mathbf{g}$ to vary freely and compare both methods introduced
in this paper to estimating these functions. We simulate each system using one
latent force and each $\boldsymbol{\beta}_r, r=0, 1$ distributed
uniformly on the sphere $S^2$. Because it is no longer possible to derive the
ground truth MAP estimates we shall instead consider the `reconstruction error'
for the IVP at the MAP estimates which we define to be the $\| \cdot \|_2$
between the true samples and the result from solving the ODE with the MAP
estimates.

The results are displayed in Table \ref{tab:so3} and for the MLFM-AG model
the same general conclusions hold; the method performs very well with $\Delta t$
small, and diminishes in performance as this increases. For the MLFM-MixSA model
we again observe that when data is sparse relative to the model structure the
greater adaptive potential of this method allows for more accurate results, we
also observe a diminishing benefit to increasing the approximation order.

\begin{table}[h]
  \caption{Reconstruction error solving the ODE
    \eqref{eq:sim_example_2} on the interval $[0, 6]$
    using the MAP estimates obtained via the MLFM-AG and the
    MLFM-MixSA methods, with two mixture components
    }
  \label{tab:so3}  
  \begin{center}
    \begin{tabular}{ccccc}
      $\Delta t$ & MLFM-AG &
      \multicolumn{3}{c}{MLFM-MixSA} \\
      & & $M=3$ & $M=5$ & $M=7$\\
      \hline \\
      0.50 & 0.110 & 0.487 & 0.212 & 0.167 \\      
      0.75 & 0.252 & 0.611 & 0.276 & 0.233 \\      
      1.00 & 0.419 & 0.570 & 0.410 & 0.355
    \end{tabular}
  \end{center}
\end{table}

\section{APPLICATION: MOCAP DATASET}
\label{sec:mocap}

Motion capture data for human poses typically consists of a representative
`skeleton', the bone segments of which are given by an orientation
vector in a local reference frame. The motion series is then given by
rotations of these initial configurations, typically recorded as Euler angles.
These angles may be represented by an equivalent unit quaternion \cite{whittaker}
and so, up to an antipodal equivalence, identified with the
sphere $S^{3} \subset \mathbb{R}^4$. 

\begin{figure}[h!]
  \centering
  \begin{subfigure}[b]{.23\textwidth}
    \includegraphics[width=\textwidth]{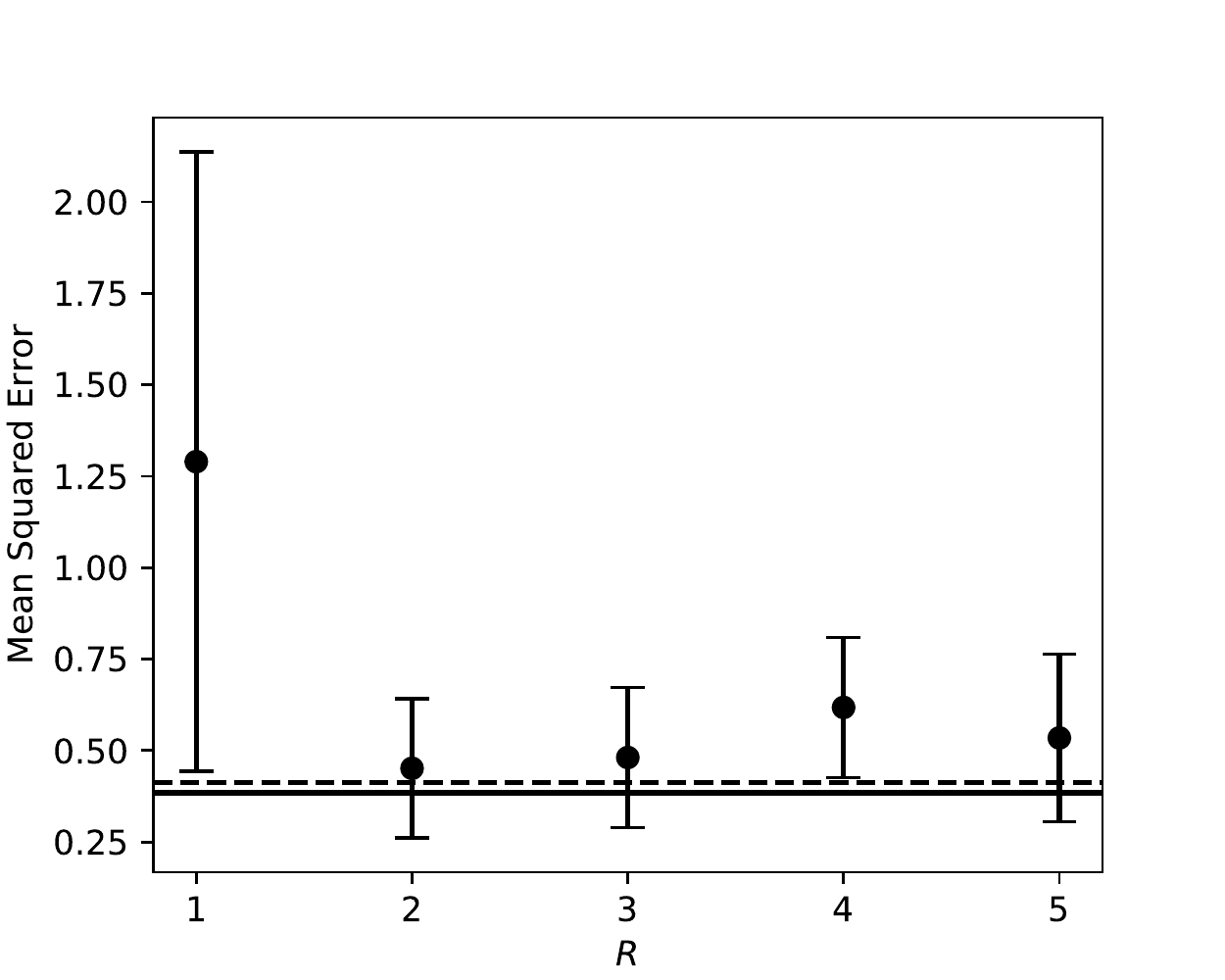}
    \subcaption{Left humerus}
  \end{subfigure}
  \begin{subfigure}[b]{.23\textwidth}
    \includegraphics[width=\textwidth]{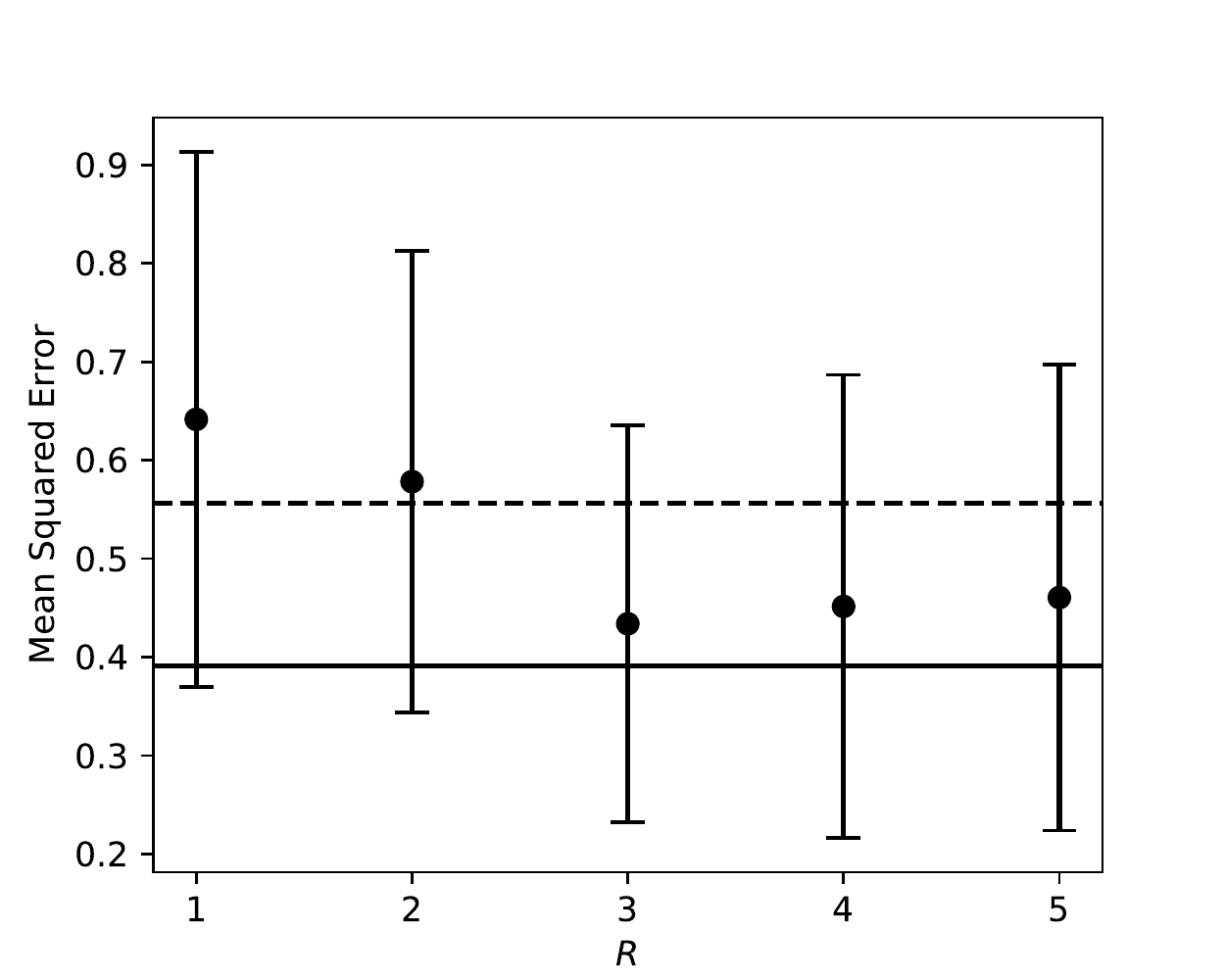}
    \subcaption{Right humerus}
  \end{subfigure}
  \begin{subfigure}[b]{.23\textwidth}
    \includegraphics[width=\textwidth]{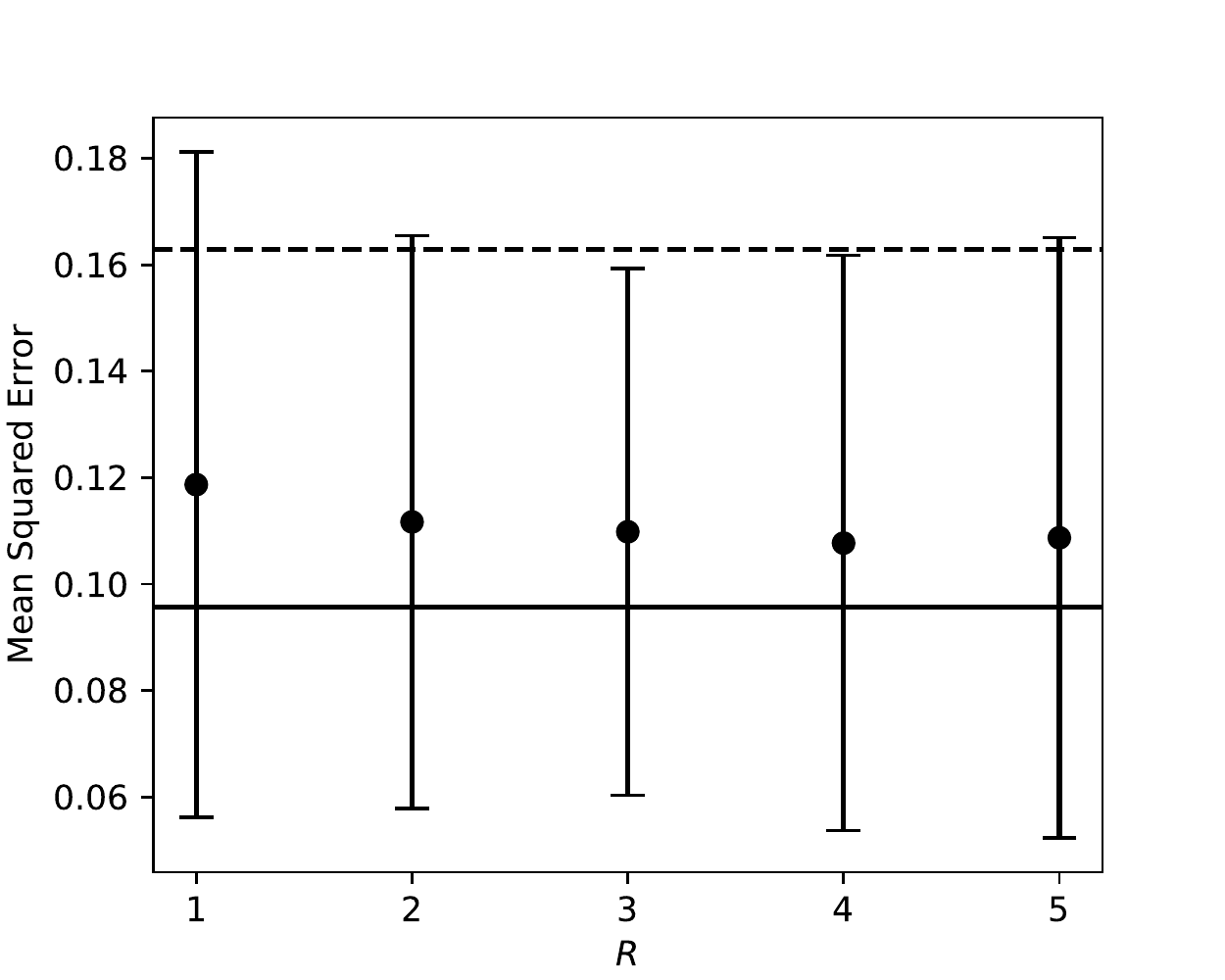}
    \subcaption{Left femur}
  \end{subfigure}
  \begin{subfigure}[b]{.23\textwidth}
    \includegraphics[width=\textwidth]{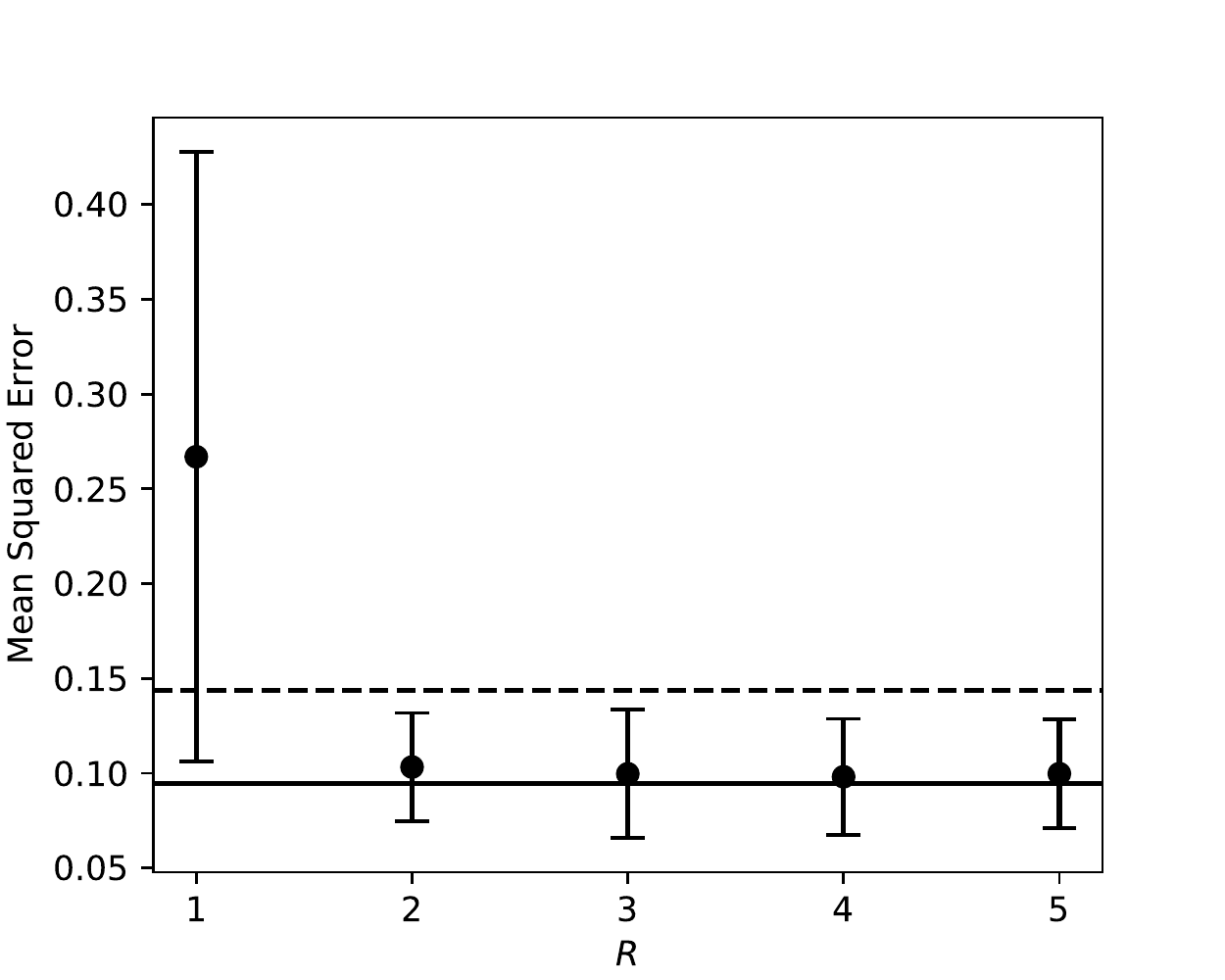}
    \subcaption{Right femur}
  \end{subfigure}
  \caption{
    Cross validation estimates of the prediction error. The error bars
    give the $\pm$ one standard deviation of the prediction error for
    each joint with $R$ latent forces fitting the MLFM with only data
    for that joint. The horizontal `\dashed' line gives
    the prediction error using the combined model with $3$ forces
    and `\full' gives the result using $4$ forces.}
  \label{fig:cv}
\end{figure}

For this experiment we first train the MLFM for the marginal time series
of each joint segment with joint specific latent variables. We use the
data from motions 1--5 of subject $64$ from the Carnegie Mellon mocap
dataset.\footnote{The
  CMU Graphics Lab Motion Capture Database was
  created with funding from NSF EIA-0196217 and is available at
  \url{http://mocap.cs.cmu.edu}} We estimate the prediction error using
leave-one-out cross validation, the results of four joints are displayed
in Figure \ref{fig:cv} which show that typically at two latent forces
are sufficient.

For the quaternion valued time series of each joint we have dimension
$K=4$, and accurate reconstructions requires $R\geq2$, as a latent variable
model this is a very modest dimension reduction. To realise the full
benefit of the latent variable modelling approach we now consider using the
product topology structure of the MLFM as described in Section \ref{sec:lfm}
by allowing each of the joints to share information through the common GPs.
The results of fitting the product MLFM with $R \in \{3, 4\}$ forces are
also displayed in Figure \ref{fig:cv}. The prediction error for the multiple
joint MLFM is given by the horizontal lines, and by sharing information
between the set of common forces we observe that the point estimates of the
model with $4$ latent forces outperforms the marginal joint prediction errors.
The product MLFM structure therefore allows not only for a latent variable
dimensionality reduction, but also the sharing of information between the joints
leads to improved generalisation and a better predictive performance.

\begin{figure*}[t!]
  \centering
  \begin{subfigure}[b]{.3\textwidth}
    \centering
    \includegraphics[width=\textwidth]{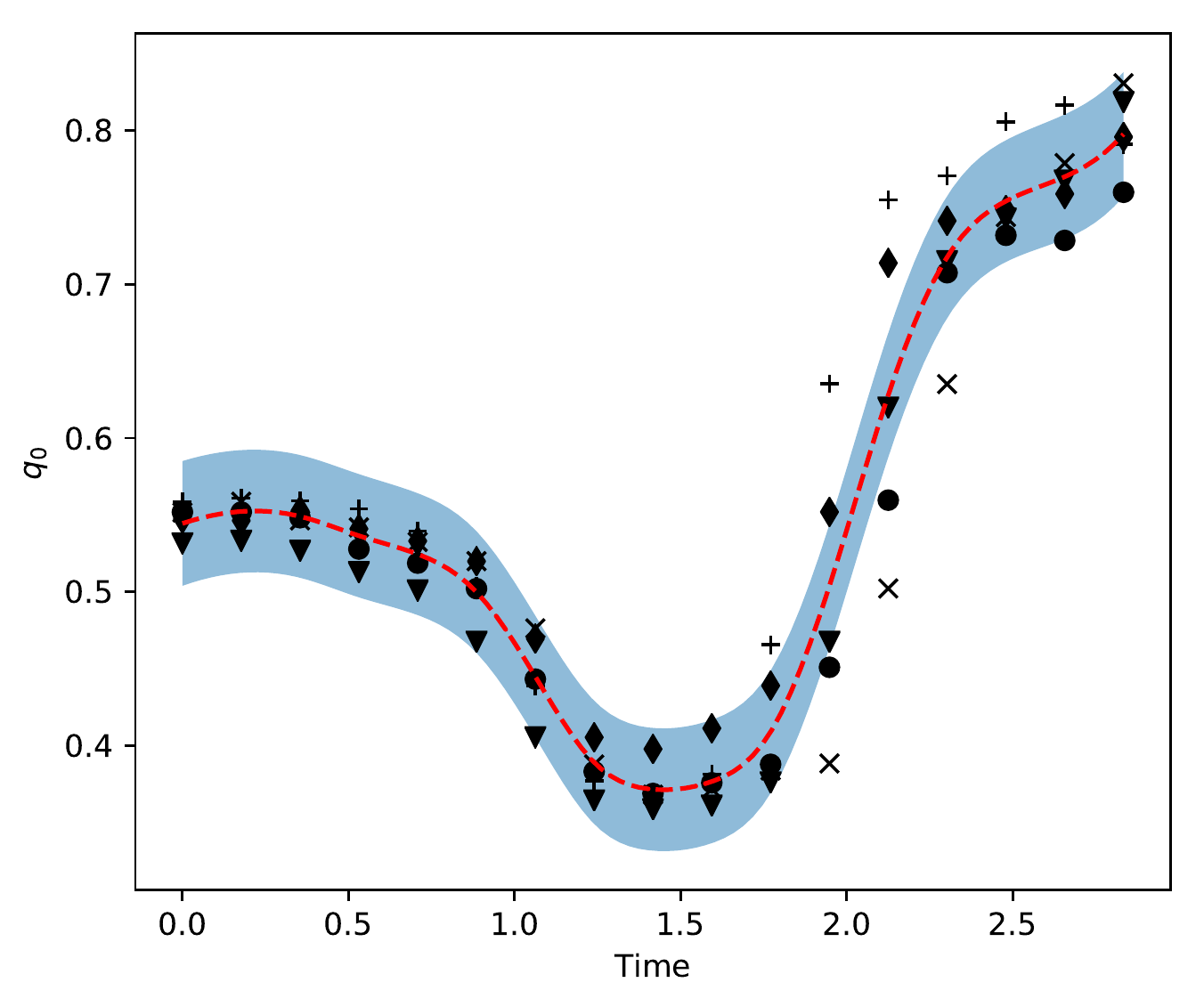}
    \caption{Left humerus, $q_0$}
  \end{subfigure}  
  \begin{subfigure}[b]{.3\textwidth}
    \centering
    \includegraphics[width=\textwidth]{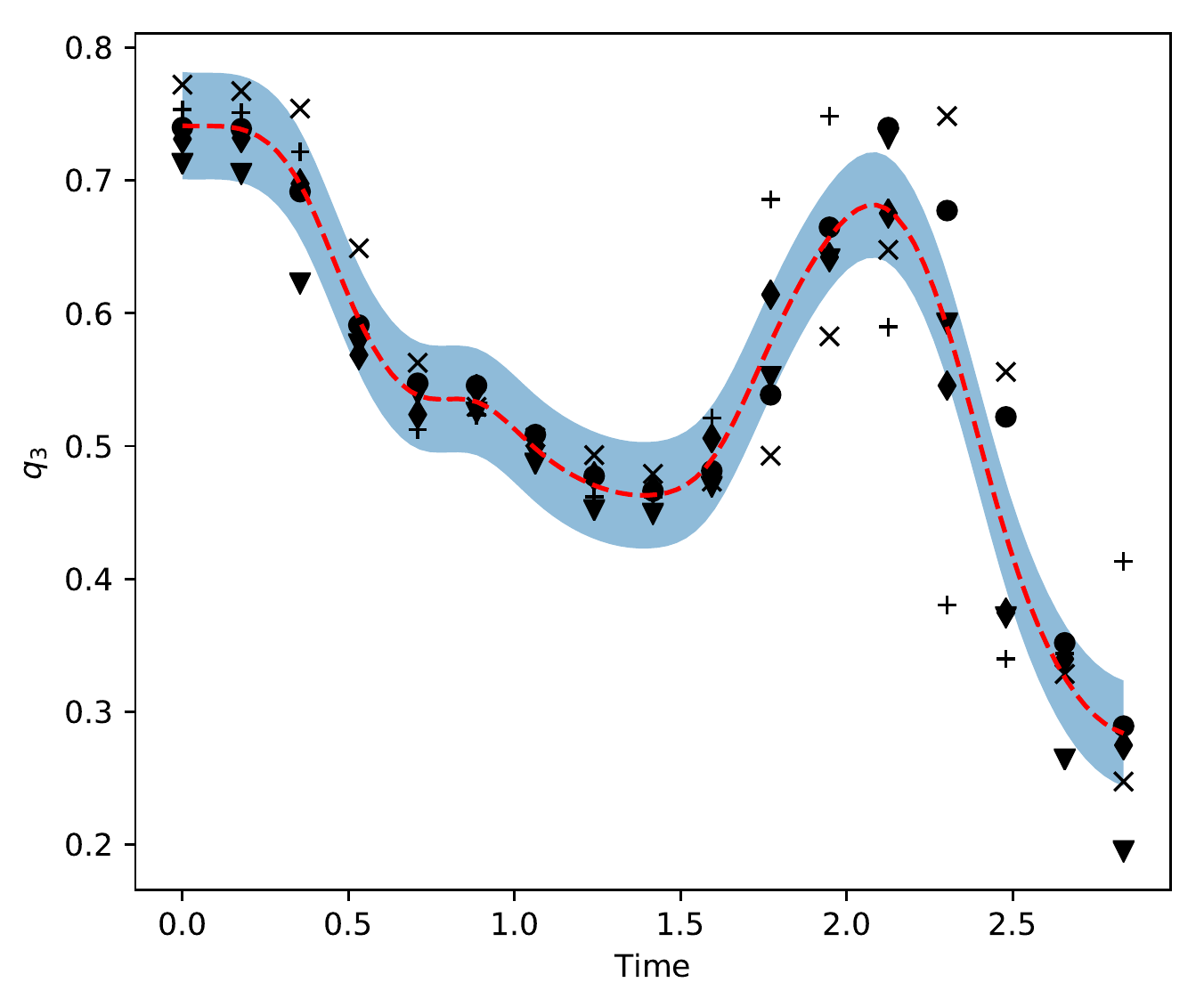}
    \caption{Left humerus, $q_3$}
  \end{subfigure}
  \begin{subfigure}[b]{.3\textwidth}
    \centering
    \includegraphics[width=\textwidth]{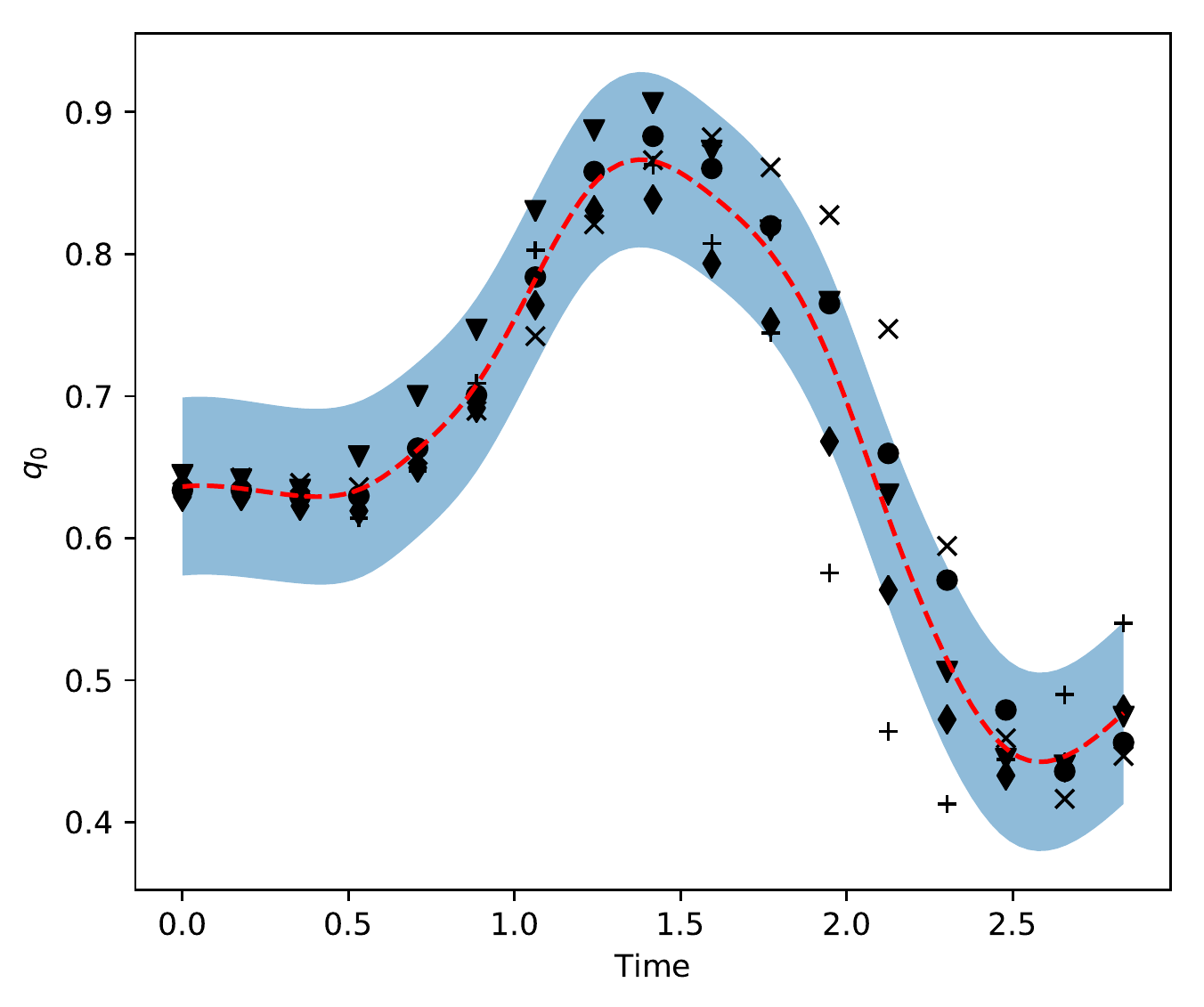}
    \caption{Right humerus, $q_0$}
  \end{subfigure}
  \centering
  \begin{subfigure}[b]{.3\textwidth}
    \centering
    \includegraphics[width=\textwidth]{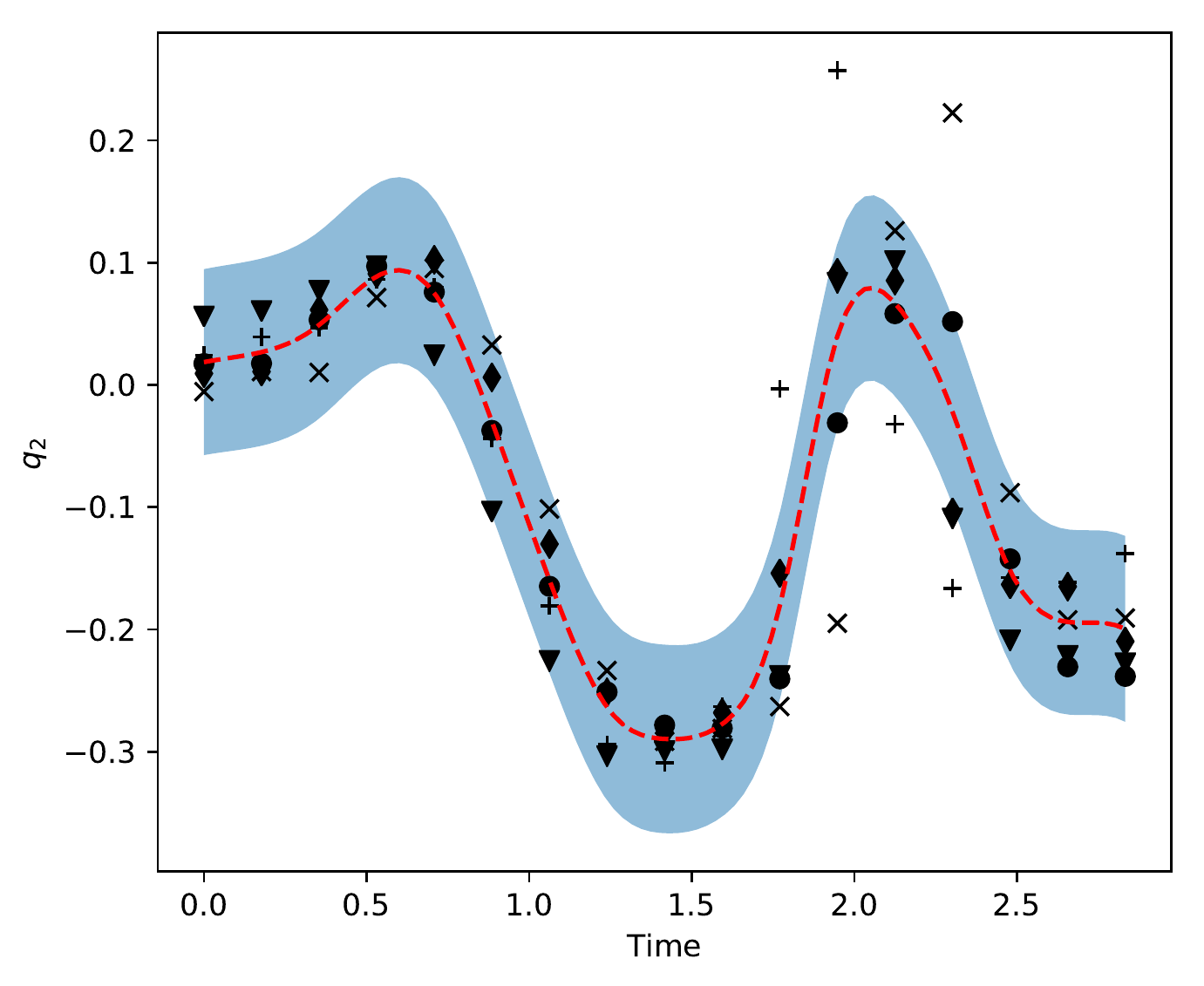}
    \caption{Right humerus, $q_2$}
  \end{subfigure}  
  \begin{subfigure}[b]{.3\textwidth}
    \centering
    \includegraphics[width=\textwidth]{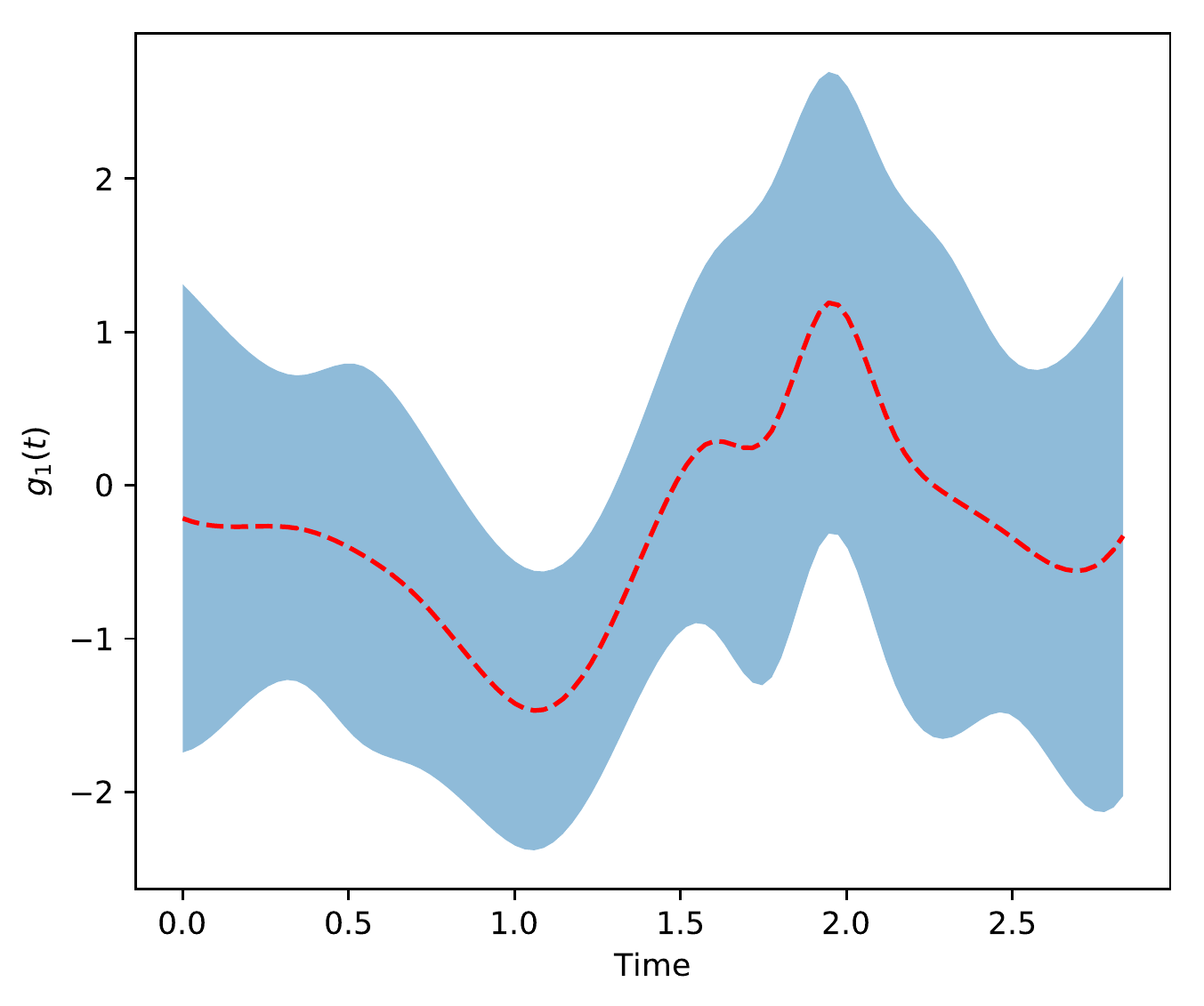}
    \caption{Right humerus, latent force 1}
  \end{subfigure}
  \begin{subfigure}[b]{.3\textwidth}
    \centering
    \includegraphics[width=\textwidth]{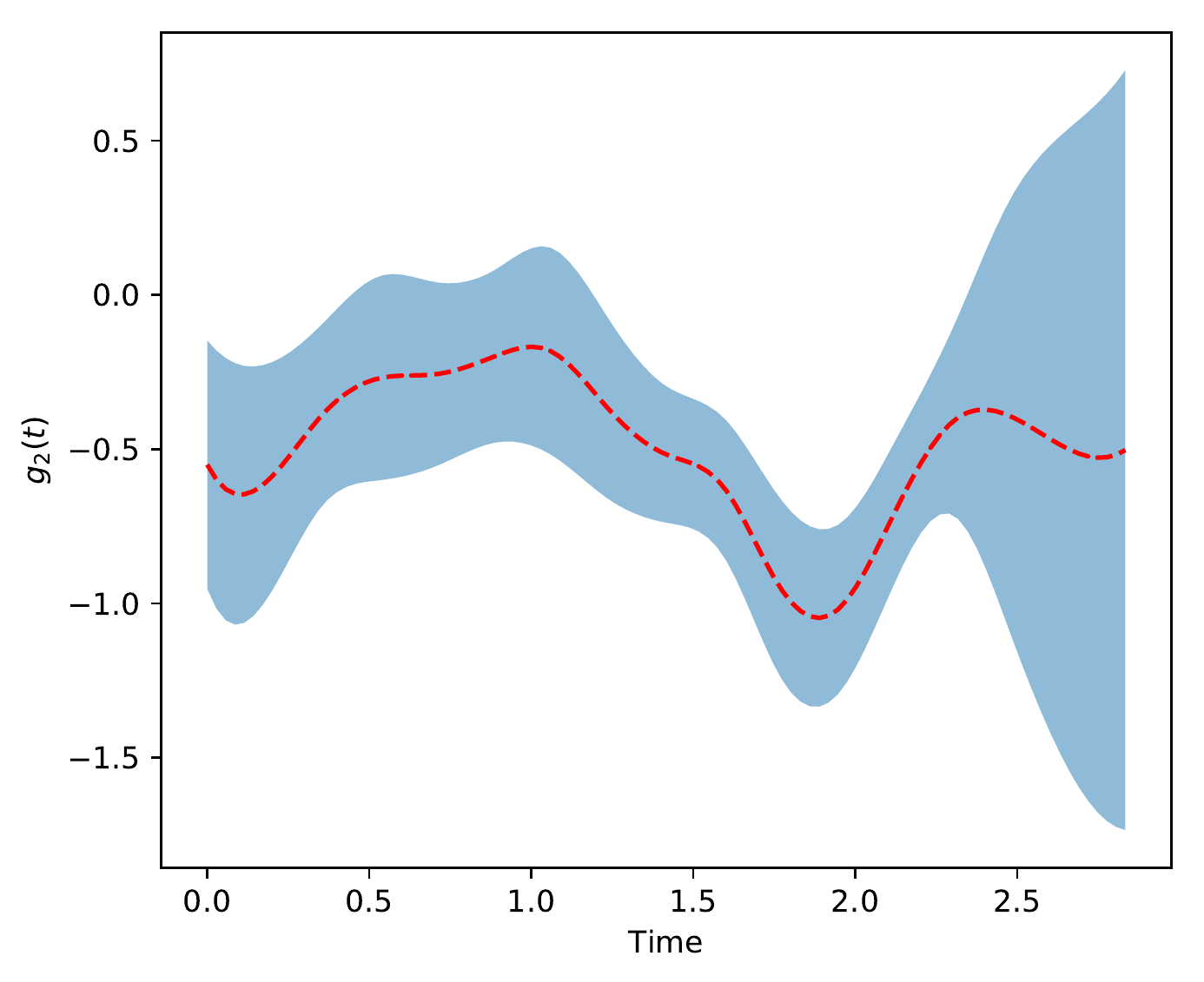}
    \caption{Multiple joint, latent force 2}
  \end{subfigure}
  \caption{Inferred motions and latent forces for subject 64 from motions
    1--5.
    }
\end{figure*}

\section{DISCUSSION}
We have described the MLFM, a semi-parameteric modelling approach for
dynamic systems allowing for strong geometric constraints to be combined
with flexible GP terms. The models provides a useful framework for modeling
trajectories on distinct manifolds sharing information through a common set
of latent GPs, and our application to motion capture data in Section
\ref{sec:mocap} indcates that this process of information sharing leads to
better predictive performance than the marginal models. To conduct inference
in this class of models we have introduced two methods of constructing
approximate conditional distributions. The first approximation was based on
the use of adaptive gradient matching methods, and the second motivated by
the method of successive approximations.

The MLFM-AG method is to be preferred in terms of computational efficiency,
however while our simulation studies presented in Section \ref{sec:sim} suggest
this method performs very well on densely sampled data, this performance
deteriorates when the data is sparse relative to the model complexity. In
particular the GP interpolants on which the model are constructred are
unable to capture the proper manifold structure over longer distances
in the data space, Figure \ref{fig:ag_S1_a}. In connection with this it
should be emphasised that any concerns of this type in the linear setting
will also manifest in the more general nonlinear setting, and therefore
embedding known structure and preserved quantities into gradient matching
methods seems an important area of future research.

The construction of the MLFM-MixSA involved discrete tuning parameters
in the truncation order, $M$, and the number of mixture components.
The method is computationally more burdensome, but the simulation
studies suggest this cost may be necessary to achieve accurate inference.
It is possible to intepret the Picard iterations \eqref{eq:picard}
as a linear dynamic system, with the approximation order as the temporal
variable. Future research may demonstrate that this may be used to motivate
an efficient approximation to the density \eqref{eq:local_density}, also
allowing for variational approaches.

One of the attractive features of the MLFM is that nonlinear changes
in the model geometry become dimensionality changes in an associated
vector space. This suggests a framework for carrying out the process of
latent manifold discovery by transforming the original problem to one in
a vector space setting. In future work we aim to apply this method to
learn topologically constrained latent variable models, \cite{Urtasun:2008},
in constrast to the a priori assumption of a known geometry in this work.

\subsubsection*{Acknowledgements}

Daniel J. Tait is supported by an EPSRC studentship.

%\begin{thebibliography}{}
\bibliography{./refs}

\begin{thebibliography}{}

\bibitem[Alvarez et~al., 2009]{alvarez}
Alvarez, M., Luengo, D., and Lawrence, N. (2009).
\newblock Latent force models.
\newblock In van Dyk, D. and Welling, M., editors, {\em Proceedings of the
  Twelth International Conference on Artificial Intelligence and Statistics},
  volume~5 of {\em Proceedings of Machine Learning Research}, pages 9--16,
  Hilton Clearwater Beach Resort, Clearwater Beach, Florida USA. PMLR.

\bibitem[Calderhead et~al., 2009]{calderhead}
Calderhead, B., Girolami, M., and Lawrence, N.~D. (2009).
\newblock Accelerating bayesian inference over nonlinear differential equations
  with {G}aussian processes.
\newblock In Koller, D., Schuurmans, D., Bengio, Y., and Bottou, L., editors,
  {\em Advances in Neural Information Processing Systems 21}, pages 217--224.
  Curran Associates, Inc.

\bibitem[Dondelinger et~al., 2013]{donde}
Dondelinger, F., Husmeier, D., Rogers, S., and Filippone, M. (2013).
\newblock Ode parameter inference using adaptive gradient matching with
  {G}aussian processes.
\newblock In Carvalho, C.~M. and Ravikumar, P., editors, {\em Proceedings of
  the Sixteenth International Conference on Artificial Intelligence and
  Statistics}, volume~31 of {\em Proceedings of Machine Learning Research},
  pages 216--228, Scottsdale, Arizona, USA. PMLR.

\bibitem[Hall, 2015]{hall}
Hall, B.~C. (2015).
\newblock {\em Lie Groups, Lie Algebras and Representations}.
\newblock Graduate Texts in Mathematics. Springer.

\bibitem[Iserles and N{\o}rsett, 1999]{iserles}
Iserles, A. and N{\o}rsett, S.~P. (1999).
\newblock On the solution of linear differential equations in {L}ie groups.
\newblock {\em Philosophical Transactions: Mathematical, Physical and
  Engineering Sciences}, 357(1754):983--1019.

\bibitem[Solak et~al., 2003]{solak}
Solak, E., Murray-smith, R., Leithead, W.~E., Leith, D.~J., and Rasmussen,
  C.~E. (2003).
\newblock Derivative observations in {G}aussian process models of dynamic
  systems.
\newblock In Becker, S., Thrun, S., and Obermayer, K., editors, {\em Advances
  in Neural Information Processing Systems 15}, pages 1057--1064. MIT Press.

\bibitem[{Tait} and {Worton}, 2018]{tait}
{Tait}, D.~J. and {Worton}, B.~J. (2018).
\newblock {Multiplicative latent force models}.
\newblock {\em ArXiv e-prints}, page arXiv:1811.00423.

\bibitem[Urtasun et~al., 2008]{Urtasun:2008}
Urtasun, R., Fleet, D.~J., Geiger, A., Popovi\'{c}, J., Darrell, T.~J., and
  Lawrence, N.~D. (2008).
\newblock Topologically-constrained latent variable models.
\newblock In {\em Proceedings of the 25th International Conference on Machine
  Learning}, ICML '08, pages 1080--1087, New York, NY, USA. ACM.

\bibitem[van Kampen, 2007]{vankampen}
van Kampen, N.~G. (2007).
\newblock {\em Stochastic Processes in Physics and Chemistry}.
\newblock Elsevier, 3 edition.

\bibitem[Whittaker and McCrae, 1988]{whittaker}
Whittaker, E.~T. and McCrae, S.~W. (1988).
\newblock {\em A Treatise on the Analytical Dynamics of Particles and Rigid
  Bodies}.
\newblock Cambridge Mathematical Library. Cambridge University Press.

\end{thebibliography}
%\end{thebibliography}
%\begin{thebibliography}{}
%\setlength{\itemindent}{-\leftmargin}
%\makeatletter\renewcommand{\@biblabel}[1]{}\makeatother
%
%\bibitem{alvarez} D.~Alvarez, D.~Luengo, and N.~D.~Lawrence (2009).
%  \newblock Latent force models.
%  \newblock In D. van Dyk, M. Welling (eds.),
%  \textit{Proceedings of the Twelth International Conference on Artifical Intelligence
%    and Statistics}
%  Clearwater Beach, Florida
%
%\bibitem{} J.~Alspector, B.~Gupta, and R.~B.~Allen (1989).
%    \newblock Performance of a stochastic learning microchip.
%    \newblock In D. S. Touretzky (ed.),
%    \textit{Advances in Neural Information Processing Systems 1}, 748--760.
%    San Mateo, Calif.: Morgan Kaufmann.
%
%\bibitem{} F.~Rosenblatt (1962).
%    \newblock \textit{Principles of Neurodynamics.}
%    \newblock Washington, D.C.: Spartan Books.
%
%\bibitem{} G.~Tesauro (1989).
%    \newblock Neurogammon wins computer Olympiad.
%    \newblock \textit{Neural Computation} \textbf{1}(3):321--323.
%
%\bibitem{tait} D.~J.~Tait, and B.~J.~Worton (2018).
%  \newblock Multiplicative Latent Force Models.
%
%  @article{tait,
%    author={D.~J.~Tait},
%    year={2018},
%    title={Multiplicative latent force models},
%    }
%\end{thebibliography}
\end{document}